\documentclass[runningheads]{llncs}
\usepackage{esvect}
\usepackage[T1]{fontenc}
\usepackage[misc]{ifsym}

\usepackage{amsmath}
\usepackage{amssymb} 
\usepackage{times}
\usepackage{soul}
\usepackage{url}
\usepackage{subfigure} 
\usepackage[misc]{ifsym}
\usepackage[utf8]{inputenc}
\usepackage[small]{caption}
\usepackage{graphicx}
\usepackage{wrapfig}
\usepackage{booktabs}
\usepackage{xcolor}
\usepackage{algorithm}
\usepackage{algorithmic}
\usepackage[switch]{lineno}
\usepackage[misc]{ifsym}

\usepackage{mwe}
\usepackage[hidelinks]{hyperref}
\begin{document}

%\title{Recent Advances in Underwater Basket Weaving Under the Extreme Pressure of the Mariana Trench}

% If the full title of your paper is short enough to also fit in the running head, you can omit the abbreviated paper title here. You can check as follows: if you comment out the \titlerunning line, something will appear in the header of all odd-numbered pages of your PDF from page 3 onward. This something is either the full title (in which case all is well), or the error message "Title Suppressed Due to Excessive Length". If this error message appears, you're going to want to provide an abbreviated title within the \titlerunning command, because if you won't do it, Springer will do it for you.

%N.B.: Author information (both in the \author{} and \authorrunning{} command) should only be present in the Camera-Ready Version of your paper. The version that you initially submit for review, ought to be double-blind. So, when initially submitting your paper, use:
%\author{Author information scrubbed for double-blind reviewing}
%\author{Andr\'e Lauren Benjamin\inst{1} \and
%Calvin Cordozar Broadus Jr.\inst{2,3} \corr \and
%Antwan Andr\'e Patton\inst{1}\orcidID{0000-1111-2222-3333}}
% You may leave out the orcidID information, if you want to.
% Use \corr to indicate the corresponding author. Note the spacing around the \corr command. Only one author can be the corresponding author.

\title{Rethinking Graph Domain Adaptation: A Spectral Contrastive Perspective}

%\titlerunning{Rethinking Graph Domain Adaptation: A Spectral Contrastive Perspective}
% Single author syntax
\author{
Haoyu Zhang\thanks{Equal contribution.}$^1$
\and
Yuxuan Cheng$^{*2}$\and
Wenqi Fan$^3$\and
Yulong Chen$^{*1}$\and
Yifan Zhang\Letter\thanks{Corresponding author. \email{yifan.zhang@cityu-dg.edu.cn}}$^1$\\}
%\affiliations

\institute{
City University of Hong Kong, Hong Kong, China\and
Huazhong Agricultural University, Wuhan, China \and
The Hong Kong Polytechnic University, Hong Kong, China
\email{
\{haoyu.zhang,yifan.zhang\}@cityu-dg.edu.cn,
hxwxss@webmail.hzau.edu.cn,
wenqifan03@gmail.com,
galahadcyl@gmail.com
}
}

\tocauthor{Haoyu Zhang,Yuxuan Cheng,Wenqi Fan,Yulong Chen,Yifan Zhang} % Replace with actual author names
\toctitle{Rethinking Graph Domain Adaptation: A Spectral Contrastive Perspective}
\maketitle 

\begin{abstract}
Graph neural networks (GNNs) have achieved remarkable success in various domains, yet they often struggle with domain adaptation due to significant structural distribution shifts and insufficient exploration of transferable patterns. 
% Traditional approaches face several limitations: they perform feature alignment directly in the spatial domain without considering scale-dependent domain shifts, treat all structural patterns equally during adaptation, and lack theoretical guarantees.
One of the main reasons behind this is that traditional approaches do not treat global and local patterns discriminatingly so that some local details in the graph may be violated after multi-layer GNN.
Our key insight is that domain shifts can be better understood through spectral analysis, where % different frequency components carry distinct structural information. Specifically, 
low-frequency components often encode domain-invariant global patterns, and high-frequency components capture domain-specific local details. As such, we propose FracNet (\underline{\textbf{Fr}}equency \underline{\textbf{A}}ware \underline{\textbf{C}}ontrastive Graph \underline{\textbf{Net}}work) with two synergic modules to decompose the original graph into high-frequency and low-frequency components and perform frequency-aware domain adaption.
Moreover, the blurring boundary problem of domain adaptation is improved by integrating with a contrastive learning framework.
Besides the practical implication, we also provide rigorous theoretical proof to demonstrate the superiority of FracNet.
Extensive experiments further demonstrate significant improvements over state-of-the-art approaches.

\end{abstract}

\section{Introduction}
% 第一段简单介绍一下图结构是很重要的，有各种应用
% 第二段大概介绍一下domain adaptation在图领域的重要性
% 第三段介绍domain adaptation的challenge
% 第四段这个motivation example
% 第五段讲这个motivation example给我们的insight（就是你要做的这个频率分解的原因）
% 第六段提出我们的方法
% 然后开始列contribution

% \begin{figure}[t]
%     \centering
%     \includegraphics[width=0.5\textwidth]{figures/node_similarity.png}
%     \caption{Node similarities between molecular pairs.
%     }
%    \label{fig:nodsim}
% \end{figure}

% 第一段：图结构数据的重要性
Graph-structured data has become increasingly ubiquitous across various domains, from social networks to molecular structures~\cite{ju2024comprehensive,jiang2023uncertainty,zhou2022graph}. The ability to effectively analyze and understand these complex graph structures is crucial for numerous applications, including drug discovery, protein structure analysis, and material science. Graph Neural Networks (GNNs) have emerged as powerful tools for learning representations from such structured data, demonstrating exceptional capabilities in capturing complex topological patterns~\cite{zhang2020deep,zhang2025strap,chen2023bridging,li2022mining}.
% 第二段：Domain Adaptation在图领域的重要性
While GNNs have achieved remarkable success in various graph-related tasks, they typically require substantial amounts of labeled training data to achieve optimal performance. However, in many real-world scenarios, particularly in specialized domains such as drug discovery~\cite{choo2023fingerprint} and materials science~\cite{butler2018machine}, obtaining labeled data can be expensive and time-consuming. This challenge has led to increasing interest in domain adaptation techniques, which aim to transfer knowledge from label-rich source domains to label-scarce target domains~\cite{cai2024graph}.

\begin{figure}[t]
    \centering
    \includegraphics[width=1\textwidth]{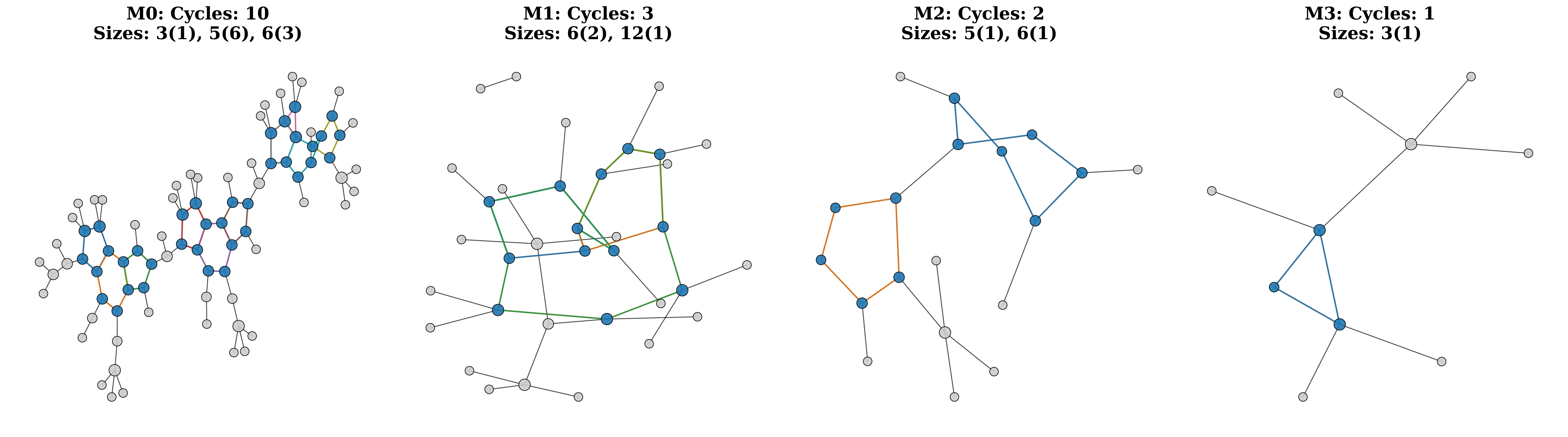}
    \caption{Molecular topological structures of four compounds.
    }
   \label{fig:120}
\end{figure}

\begin{figure}[t]
    \centering
    \includegraphics[width=1\textwidth]{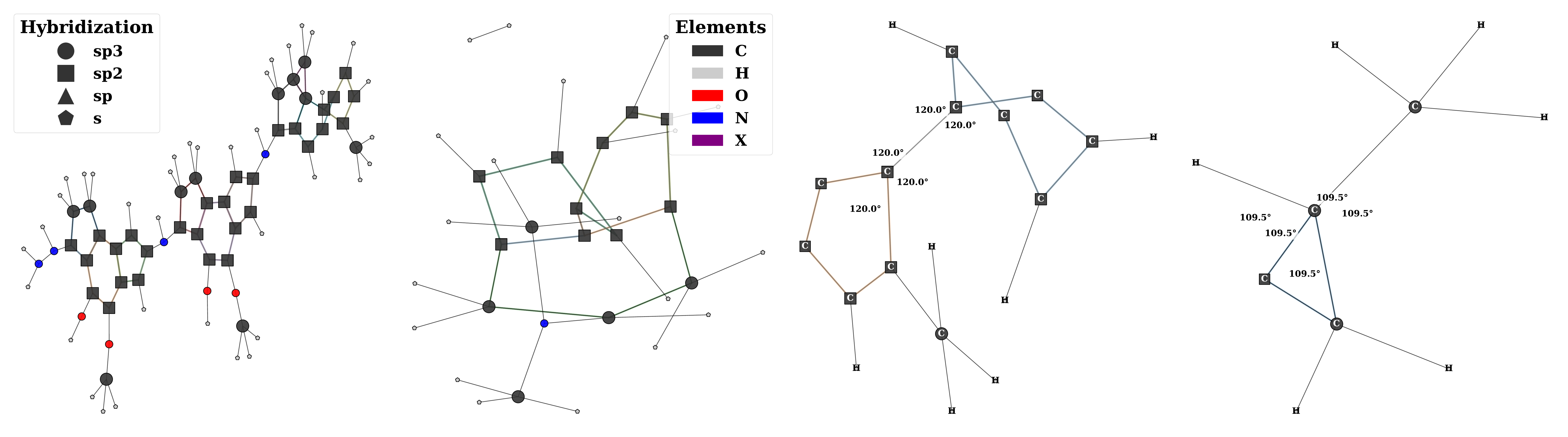}
    \caption{Molecular properties of four compounds.
    }
   \label{fig:mol}
\end{figure}

% 第三段：Domain Adaptation的挑战
However, existing efforts treat the graph as a holistic entity without distinguishing its properties in global and local patterns~\cite{liu2024rethinking}, which may lead to inaccurate performance due to overlooking crucial local structural features. 
For example, consider representative molecules from different domains of the Mutagenicity dataset, as shown in Figure~\ref{fig:mol}. While M0 contains multiple cyclic structures (10 cycles in total, including 6 five-membered rings and 3 six-membered rings), M1 exhibits fewer but larger cycles (3 cycles, including 2 six-membered rings and 1 twelve-membered ring). These structural differences are further emphasized by their hybridization states and bond angles. M0 is rich in sp$^2$ hybridized carbons forming planar geometries with 120° bond angles, characteristic of aromatic rings with delocalized $\pi$-electrons. 
In contrast, M3 contains predominantly sp$^3$ hybridized carbons with tetrahedral geometry and 109.5° bond angles, forming more flexible single bonds. 
These distinct local structural motifs, particularly the type and number of cyclic systems, significantly influence molecular reactivity, electron distribution, and biological properties. 
Traditional GNNs, primarily focusing on topological connectivity, often fail to effectively capture these critical local geometric features that determine molecular behavior across different domains.

This observation naturally introduces our key insight: domain shifts in graph data can be better understood and addressed through spectral analysis. 
When projecting graphs into the frequency domain, different structural patterns—such as the sp$^2$-rich aromatic systems in M0 versus the sp$^3$-dominated structures in M3—manifest as distinct frequency components with varying degrees of transferability across domains. 
Specifically, low-frequency components often correspond to groups of strongly connected nodes with similar features, capturing \textbf{global}, potentially transferable patterns like the basic carbon scaffolds common across domains. 
In contrast, high-frequency components reflect rapid variations in node features between neighborhoods, often representing domain-specific \textbf{local} details such as the specific ring sizes, hybridization states, and bond angles that differentiate molecular domains. 
This frequency-domain perspective provides a principled way to understand and address domain shifts in molecular graph datasets.

% 第六段：提出方法
Based on these observations, we propose FracNet, a new framework that combines frequency decomposition with contrastive learning for better domain adaptation. Our method first breaks down graph structures into high- and low-frequency parts, which helps us understand different structural patterns in molecular graphs. 
The low-frequency parts show us the overall structure that tends to be similar across different domains, while the high-frequency parts capture detailed local features that might be domain-specific.
Given the decomposed components, we then improve the conventional Maximum Mean Discrepancy (MMD), a popularly used method for domain alignment, from the following two perspectives: (1) mitigate the problem of blurring class boundaries in binary classification tasks~\cite{chen2019transferability}; 
(2) extend it into the frequency domain, considering multiple frequency components.
To this end, we use contrastive learning~\cite{he2020momentum} to maintain clear class separation while aligning domains to achieve the first enhancement. And the second one is addressed by designing a new kernel to integrate the components of all frequencies. This approach helps us achieve better transfer learning results, especially for molecular classification tasks.
% Contributions
Our main contributions can be summarized as follows:
\begin{itemize}
    \item We introduce a new method that uses frequency decomposition to analyze molecular structures. This approach helps us better understand and transfer knowledge between different molecular domains by separating global patterns from local details. The method is especially useful for molecular property prediction tasks where overall structure and local features both matter.
    
    \item We propose a novel combination of contrastive learning and MMD alignment that helps solve the negative transfer problem in binary classification tasks. This combination maintains clear class boundaries while aligning different domains, leading to better classification results.

    \item We design a frequency-aware kernel that further enhances MMD-based domain alignment by separating and aligning molecular features at different frequencies, significantly outperforming traditional Gaussian kernels in capturing both global structural similarities and local molecular patterns.
    
    \item Our results demonstrate that understanding molecular structures through frequency decomposition and using contrastive learning can significantly improve domain adaptation performance.
\end{itemize}

\section{Related Work}

\subsection{Domain Alignment}
Domain alignment has emerged as a fundamental paradigm in transfer learning, particularly for scenarios with limited labeled data in the target domain~\cite{liu2023adversarial,wang2023correspondence}. Recent advances have focused on developing more sophisticated alignment strategies to handle complex domain shifts.~\cite{liu2023joint} propose an adversarial framework with contrastive learning, while~\cite{sun2023rethinking} present a unified framework that disentangles and aligns different aspects of the data distribution. In the context of structural data,~\cite{hu2024domain} explore the synergy between internal feature exploration and external domain alignment. However, these methods typically rely on empirical objectives without rigorous theoretical guarantees on the optimality of alignment. Our work addresses this limitation by establishing theoretical connections between contrastive learning and Maximum Mean Discrepancy (MMD), providing provable bounds on alignment quality through frequency decomposition.

\subsection{Graph Spectral Processing}
Graph spectral processing has revolutionized the analysis of graph-structured data by leveraging frequency domain representations~\cite{bo2024graph,jiang2023incomplete,zhang2024infinite,jiang2024ragraph,zhang2023spectral}. ~\cite{bo2023specformer} introduce Specformer by incorporating spectral graph processing with transformers, while~\cite{rattan2023weisfeiler} bridge the gap between Weisfeiler-Leman algorithms and graph spectra. Recent work has focused on enhancing spectral methods' robustness, with~\cite{gao2023addressing} addressing heterophily through spectral analysis. While these methods demonstrate the effectiveness of spectral processing, they lack theoretical guarantees on the optimality of frequency decomposition in domain adaptation. Our framework addresses this limitation by proving that the proposed spectral decomposition achieves tighter mutual information bounds, with explicit guarantees on both local and global structural alignment.

\section{Methodology}
\begin{figure}
    \centering
    \includegraphics[width=1\linewidth]{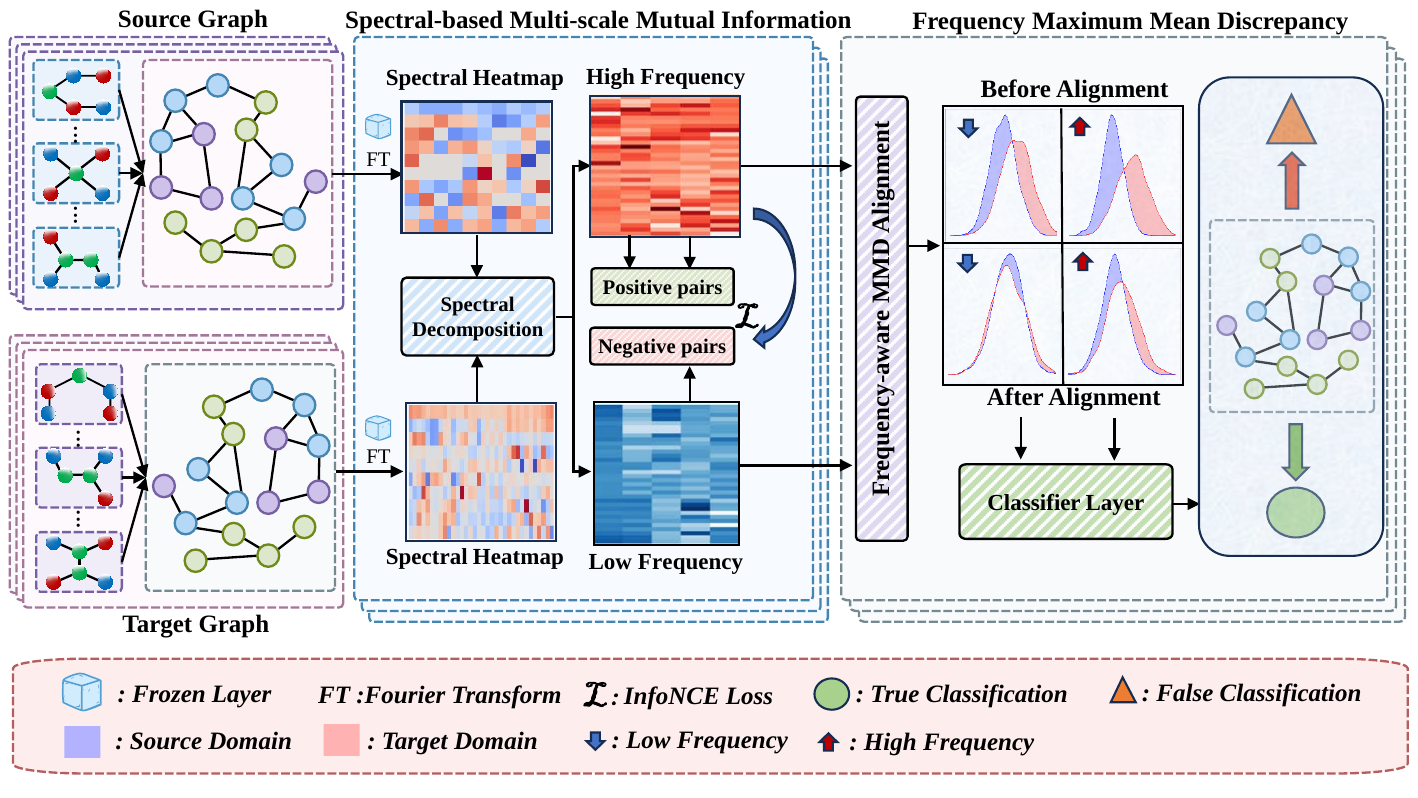}
    \caption{Framework overview of FracNet. The model decomposes source and target graphs into high/low frequency components via Fourier Transform, followed by dual-stream processing with SMMI for contrastive learning and FMMD for domain alignment.}
    \label{fig:pipline}
\end{figure}
% The overview of the proposed FracNet framework for unsupervised domain adaptive graph classification is illustrated in Figure 1. The core of our FracNet is to provide frequency-aware learning to bridge domain gaps using spectral analysis. Theoretically, we utilize a spectral-guided maximum mutual information (SMMI) mechanism to capture frequency-specific features (see Section 3.2). In contrast, a frequency-aware maximum mean discrepancy (FMMD) module is adopted to align cross-domain distributions in different frequency bands (see Section 3.3), theoretically guaranteeing more precise domain alignment. Moreover, we incorporate the two modules into a holistic spectral contrastive learning framework (see Section 3.4). On the one hand, we perform frequency decomposition with provable bounds to extract complementary information from different frequency bands. On the other hand, contrastive learning with theoretical guarantees aims to maximize the mutual information between corresponding frequency components while minimizing the MMD distances between cross-domain sample pairs in the frequency domain.

% Given the remarkable capacity of contrastive learning to discover discriminative representations through instance-level discrimination while maintaining intrinsic graph topology and structural correspondence across domains, 
To achieve the domain adaptation considering both global and local patterns, we propose FracNet, a theoretically grounded spectral contrastive framework. 
The core of FracNet comprises two synergistic modules, a Spectral-guided Maximum Mutual Information (SMMI) module and a Frequency-aware Maximum Mean Discrepancy (FMMD) module, as illustrated in Figure~\ref{fig:pipline}.
Specifically, SMMI leverages spectral decomposition to disentangle graph signals into complementary frequency bands, enabling the model to capture both global topological invariants in low-frequency components and fine-grained structural variations in high-frequency components.
FMMD proposed a new kernel to implement domain alignment in the frequency domain.
Moreover, as FMMD is designed based on the most famous domain adaptation framework MMD, we integrate a contrastive framework in SMMI to further contribute to improving the blurring boundary problem of conventional MMD.

% Given the remarkable capacity of discovering and maintaining discriminative representations across domains, we exploit contrastive learning as the base of the FMMD module to achieve cross-domain alignment at different spectral scales, ensuring optimal knowledge transfer through frequency-dependent alignment strategies with theoretical guarantees on adaptation performance. 

% facilitates precise cross-domain alignment by explicitly minimizing distribution discrepancy at different spectral scales via contrastive learning, ensuring optimal knowledge transfer through frequency-dependent alignment strategies with theoretical guarantees on adaptation performance. 
% These two modules operate synergistically: 
% SMMI establishes frequency-specific feature representations with maximized mutual information within each domain, while FMMD optimizes the transfer of these spectral components across domains through  distribution alignment, collectively forming a unified framework that addresses the fundamental challenges in graph domain adaptation from a spectral perspective.

\subsection{Problem Formulation}
Given a graph $\mathcal{G} = (\mathcal{V}, \mathcal{E})$, where \(\mathcal{V}\) is the set of nodes and $\mathcal{E} \subseteq V \times V $ denotes the set of edges. For each graph, we denote its Laplacian matrix by \(L = D - A \in \mathbb{R}^{|\mathcal{V}| \times |\mathcal{V}|}\), where $D$ is the degree matrix and $A$ is the adjacency matrix. In our problem, we have access to a labeled source domain $D^S = {\{(G^s_i,y^s_i)\}}^{n_s}_{i=1}$ with $n_s$ samples and an unlabeled target domain $D^t = \{ G^t_j \}_{j=1}^{n_t}$  with $n_t$ samples.  $D^s$ and $D^t$ share the same label space  $y = \{1,2\}$ but with different distributions in the data space. Our objective is to train the graph classification model on both $D^s$ and $D^t$, and attain high accuracy on the test dataset of the target domain.

\subsection{Spectral-guided Maximum Mutual Information}
Domain adaptation for graph-structured data faces unique challenges due to the complex interplay between node features and topological structures. 
Traditional methods often treat graphs as holistic entities, failing to capture the multi-scale nature of graph signals. 
Graph spectral transforms~\cite{sandryhaila2013discrete} project node features onto the eigenbasis of the graph Laplacian, where the eigenvectors reflect different patterns of node relationships: low-frequency components correspond to strongly connected nodes with similar features, while high-frequency components capture nodes with dissimilar features between local neighborhoods. 
This decomposition enables us to analyze and align domain shifts at different levels of node relationships.
Although graph spectral transform has been well studied for years, integrating the decomposed components to further contribute to domain adaptation still remains challenging.
To address this challenge, we propose a novel Spectral-guided Maximum Mutual Information (SMMI) mechanism. 

% The key insight to achieve reasonable decomposition lies in filtering graph signals based on node similarity patterns. 
% When projecting signals onto the graph Laplacian eigenbasis, low-frequency components correspond to groups of strongly connected nodes with similar features, 
% while high-frequency components reflect feature differences between neighboring nodes. 
% This spectral view naturally separates the consistent and varying patterns of node relationships. 

Given source domain sample $z_s$, target domain sample $z_t$, and negative sample $z_n$, we decompose them into low-frequency and high-frequency components through graph Fourier transform while preserving the original graph structure:
\begin{equation}
    z_s = \begin{pmatrix} \lambda_l z_{s,l} \\ \lambda_gz_{s_g} \end{pmatrix}, z_t = \begin{pmatrix} \lambda_l z_{t,l} \\ \lambda_gz_{t_g} \end{pmatrix}, z_n = \begin{pmatrix} \lambda_l z_{n,l} \\ \lambda_gz_{n_g} \end{pmatrix}, 
\end{equation}
Where $\lambda_l$ and $\lambda_g$ are the weights for low-frequency and high-frequency components, respectively. Note that $z_n$ is defined here since we mainly focus on binary classification task and introduce a contrastive framework in the rest of this section.
A
For vectors $z_l = (\lambda_lz_{i,l},\lambda_gz_{i,g})^T$, where $i \in \{s,t,n\}$,  $z_{i,l}$ and $ z_{i,g}$ are orthogonal
%\footnote{Detailed proof can be found in the supplementary material} $(z^T_{i,l}z_{i,g}=0)$, then
$z_ {i}^ {T}z_ {j} = \lambda _ {l}^ {2}   z_ {i,l}^ {T} z_ {j,l}  + \lambda _ {g}^ {2}z_ {i,g}^T  z_ {j,g}  ,j  \in \{s,t,n\} $.

Based on this decomposition, we design the following contrastive learning objective to guarantee the discriminative representations for different frequencies:
\begin{equation}
%\scriptstyle
    \mathcal{L} = -\frac{1}{|P|}\sum_{(z_s,z_t)\in P}\left(\frac{Z_s^Tz_t}{\tau}-\log\left(\sum_{z_n \in N}e^{\frac{Z_s^Tz_n}{\tau}}+\sum_{z_n \in N}e^{\frac{Z_t^Tz_t}{\tau}}\right)\right)
\end{equation}
Where $P$ is the set of positive pairs and $\tau$ is the temperature parameter.
To better understand the behavior of this loss function, we introduce cosine similarity measures
$\cos(\theta_{i,j,l}) =\frac{z^T_{i,l}z_{j,l}}{\|z_{i,l}\|\|z_{j,l}\|},\ \cos(\theta_{i,j,g}) =\frac{z^T_{i,g}z_{j,g}}{\|z_{i,g}\|\|z|_{j,g}\|},\ i,j \in \{s,t,n\}$.

We assume that features are normalized, i.e., $\|z_{i,l}\| \approx \|z_{j,l}\| \approx K_l ,\|z_{i,g}\| \approx \|z_{j,g}\| \approx K_g$, and define $\tilde{\lambda}_l^2 = \lambda_l^2k_l^2 , \tilde{\lambda}_g^2 = \lambda_g^2k_g^2$. This normalization assumption is both theoretically motivated and practically beneficial. From a theoretical perspective, it ensures that the cosine similarities are well-defined and bounded. From a practical standpoint, it stabilizes training and makes the learning process more robust to numerical issues. Using these normalized features, we can rewrite our contrastive loss in a more interpretable form:
\begin{align}
\scriptsize
\begin{split}
    &\mathcal{L}_{contrast} \approx -\frac{1}{|P|} \sum_{(z_s,z_t)\in P} \Bigg( \frac{\tilde{\lambda_l^2}\cos(\theta_{s,t,l})+\tilde{\lambda_g^2}\cos(\theta_{s,t,g})}{\tau} \\ &
    -\log\Bigg(\sum_{z_n \in N}exp\Bigg(\frac{\tilde{\lambda_l^2}\cos(\theta_{s,n,l})+\tilde{\lambda_g^2}\cos(\theta_{s,n,g})}{\tau}\Bigg) 
    +\sum_{z_n \in N}exp\Bigg(\frac{\tilde{\lambda_l^2}\cos(\theta_{t,n,l})+\tilde{\lambda_g^2}\cos(\theta_{t,n,g})}{\tau}\Bigg)\Bigg)\Bigg)
\end{split}
\end{align}

To analyze this loss function rigorously, we employ Jensen’s inequality$^*$. Through algebraic manipulation, we can reorganize the terms to highlight the separate contributions of low and high-frequency components:
\begin{align}
%\scriptsize
\begin{split}
    \mathcal{L}_{SMMI} & \simeq - \frac{\tilde{\lambda_l^2}}{\tau} \left( \frac{1}{|P|} \sum_{(z_s,z_t) \in P} \cos(\theta_{s,t,l}) \right.  \quad \left. - \frac{1}{2} \mathbb{E}_{z_n \sim D_n}[\cos(\theta_{s,n,l}) + \cos(\theta_{t,n,l})] \right) \\
    & \quad - \frac{\tilde{\lambda_g^2}}{\tau} \left( \frac{1}{|P|} \sum_{(z_s,z_t) \in P} \cos(\theta_{s,t,g}) \right. \quad \left. - \frac{1}{2} \mathbb{E}_{z_n \sim D_n}[\cos(\theta_{s,n,g}) + \cos(\theta_{t,n,g})] \right)
\end{split}
\end{align}

This decomposition reveals two important properties of our SMMI mechanism: 1) The loss function naturally separates into low-frequency ($\tilde{\lambda^2_l}$) and high-frequency ($\tilde{\lambda^2_g}$) components, allowing for independent optimization of each frequency band; 2) Each frequency component maintains a balance between positive pair attraction (first term) and negative pair repulsion (second term).

\subsection{Frequency-aware Maximum Mean Discrepancy}
\label{subsec:fmmd}

Following the frequency decomposition, we propose a principled approach to measure and align the domain distributions across different frequency components. While Maximum Mean Discrepancy (MMD) has been widely adopted for distribution alignment, traditional MMD-based methods typically employ Gaussian kernels:
\begin{equation}
    k_{gaussian}(x,y)=exp(-\frac{\|x-y\|^2}{2\sigma^2})
\end{equation}
Where $\sigma$ is the bandwidth parameter. Though effective for general distribution alignment, Gaussian kernels treat all feature dimensions uniformly without considering the intrinsic spectral characteristics of graph-structured data. This limitation motivates us to design a frequency-aware kernel function. We propose a novel kernel function that explicitly measures similarities in both low and high-frequency components:
\begin{equation}
    k(x,y)=\cos(\theta_{x,y,l})+\cos(\theta_{x,y,g})
\end{equation}
where $\theta_{x,y,l}$ and $\theta_{x,y,g}$ represent the angles between samples in low- and high-frequency spaces, respectively. This design offers several key advantages over conventional Gaussian kernels: 1) Our kernel explicitly captures the angular relationships in different frequency bands, enabling more precise analysis of spectral distributional shifts; 2) Unlike Gaussian kernels that are sensitive to feature scaling, the cosine-based design is naturally invariant to the magnitude of features, making it more robust to frequency-specific variations; 3) The separation of frequency components provides clear geometric interpretation of distribution differences in the spectral domain, facilitating better understanding of domain gaps.

The proposed kernel satisfies several important theoretical properties. For random variables $X,X' \sim D$ independently sampled from distribution $D$, there exist $c_D \in \mathbb{R}$ and $\epsilon_D:\Theta \rightarrow \mathbb{R}$, such that:
\begin{equation}
    \mathbb{E}_{X,X'\sim D}[k(X,X')]=c_D+\epsilon_D(\theta)
\end{equation}
With the following properties:
\begin{enumerate}
    \item $c_D$ depends only on distribution $D$
    \item $\sup_{\theta \in \Theta}|\epsilon_D(\theta)|\leq \delta$ for some small constant $\delta > 0$
    \item $\|\nabla_{\theta \in D}(\theta)\|_2\leq \eta$ for some small constant $\eta >0$
\end{enumerate}

Furthermore, we prove that our kernel satisfies two crucial mathematical properties$^*$:
\begin{enumerate}
    \item \textbf{Lipschitz Continuity}: For any $x,y,x',y' \in X$:
    \begin{equation}
        |k(x,y)-k(x',y')|\leq L(\| x-x'\|_2+\| y-y'\|_2)
    \end{equation}
    \item \textbf{Boundedness}: For any $x,y \in X$:
    \begin{equation}
        |k(x,y)|\leq M
    \end{equation}
\end{enumerate}
Where $L$ and $M$ are positive constants.

Based on this theoretically grounded kernel design, we can formulate the MMD between source and target domains as follows:
\begin{equation}
\begin{split}
    MMD^2(D_s,D_t) &= \mathbb{E}_{x,x' \sim D_s}[k(x,x')] + \mathbb{E}_{y,y' \sim D_t}[k(y,y')] - 2\mathbb{E}_{x \sim D_s,y \sim D_t}[k(x,y)] \\
    &= c_{D_s}+c_{D_t} - \frac{2}{|P|}\sum_{(z_s,z_t) \in P}(\cos(\theta_{s,t,l})+\cos(\theta_{s,t,g}))   
\end{split}
\end{equation}
Where $c_{D_s}$ and $c_{D_t}$ are distribution-specific constants.%, and $P$ denotes the set of paired samples.

To handle negative samples effectively, we extend our analysis to measure the distributional differences between positive and negative samples:

\begin{equation}
\begin{aligned}
    MMD^2(D_s,D_n) &= c_{D_s}+c_{D_n} - 2\mathbb{E}_{z_n \sim D_n}[\cos(\theta_{s,n,l})+\cos(\theta_{s,n,g})]
    \\
    MMD^2(D_t,D_n) &= c_{D_t}+c_{D_n} - 2\mathbb{E}_{z_n \sim D_n}[\cos(\theta_{t,n,l})+\cos(\theta_{t,n,g})]
\end{aligned}
\end{equation}
Where $D_n$ represents the negative sample distribution.

Furthermore, we can get $\mathcal{L}_{FMMD}$:
\begin{equation}
\begin{split}
    \mathcal{L}_{FMMD} =& 
     -[\mathbb{E}_{z_n \sim D_n}[\cos(\theta_{s,n,l})+\cos(\theta_{s,n,g})] 
    +\mathbb{E}_{z_n \sim D_n}[\cos(\theta_{t,n,l})+\cos(\theta_{t,n,g})]]
\end{split}
\end{equation}

\subsection{Unified Spectral Contrastive Framework} A key theoretical insight of our work is that the SMMI objective and FMMD alignment are inherently connected through our frequency-aware kernel design. We get the final loss function %\footnote{Detailed proof can be found in the supplementary material}:
% \begin{equation}
% \begin{split}
$ \mathcal{L} = \mathcal{L}_{CE} + \gamma_1 \mathcal{L}_{SMMI} + \gamma_2\mathcal{L}_{FMMD} $
% \end{split}
% \end{equation}
Where  $\gamma_1$ and $\gamma_2$ are balance parameters.
This integrative approach ensures that the learned representations are both discriminative within domains and transferable across domains, while preserving the multi-scale nature of graph-structured data.
\section{Experiments}
In this section, we conduct extensive experiments to evaluate the effectiveness of our proposed FracNet framework. Our experiments aim to answer the following key questions:

\begin{itemize}
    \item \textbf{RQ1:} How does our FracNet perform compared to state-of-the-art domain adaptation methods for graph-structured data?
    
    \item \textbf{RQ2:} How do the spectral relationships between different domains (high- and low-frequency components) affect cross-domain classification accuracy?
    
    \item \textbf{RQ3:} What impact does the graph structure of molecules have on cross-domain classification performance?
    
    \item \textbf{RQ4:} How do different components and hyperparameters of our model contribute to the overall performance?
\end{itemize}
\subsection{Experimental Settings}
\textbf{Datasets.}~We conduct extensive experiments on three widely-used benchmark datasets from TUDataset~\cite{morris2020tudataset} in the setting of unsupervised domain adaptation. For convenience M, N, and P are short for Mutagenicity, NCI1, and PROTEINS, respectively. Their details are introduced as follows in Table~\ref{tab:datasets}.
\textbf{Mutagenicity}~\cite{kazius2005derivation} contains 4337 chemical compounds with corresponding Ames test data indicating their mutagenic effect. % Similar to FRANKENSTEIN. We divide them into four sub-datasets (M0, M1, M2 and M3) based on  on the edge density.
\textbf{NCI1}~\cite{wale2008comparison} consists of 4110 chemical compounds screened for activity against non-small cell lung cancer. % We divide them into four sub-datasets (N0, N1, N2 and N3) based on  on the edge density.
\textbf{PROTEINS}~\cite{dobson2003distinguishing} contains 1113 proteins where nodes represent amino acids and edges indicate spatial proximity (distance $<$ 6 Angstroms). % The task is to classify proteins as enzymes or non-enzymes. We divide them into four sub-datasets (P0, P1, P2 and P3) based on  on the edge density.
Each of these datasets is divided into different domains based on node density.
M0 represents the domain 0 of Mutagenicity dataset and the rest can be deduced by analogy.
\setlength{\tabcolsep}{0.4em}
\begin{table}[ht]
    %\small
    \centering
    \begin{tabular}{lcccc}
    \toprule
        Datasets & Graphs &Avg. Nodes &Avg. Edges & Classes \\ 
    \midrule
         Mutagenicity& 4337 &30.32&30.77&2\\
          NCI1& 4110&29.87&32.30&2\\
          PROTEINS &1113 &39.1 &72.8&2 \\ 
    \bottomrule
    \end{tabular}
    %\vspace{-0.1cm}
    \caption{Statistics of the experimental datasets.}
    \label{tab:datasets}
\vspace{0.0005cm}
\end{table}

\textbf{Baselines.}~We compare FracNet with a wide range of existing methods. These baseline methods fall into three categories: (1) Graph neural networks, e.g., GCN~\cite{kipf2016semi}, GIN~\cite{xu2018powerful}, GMT~\cite{baek2021accurate}, GAT~\cite{velickovic2017graph}, GraphSAGE ~\cite{hamilton2017inductive} and DeSGDA~\cite{wang2024degree}. These methods only use the source domain data for training and test on target domain data. 
(2) Unsupervised domain adaptation methods, e.g., CDAN~\cite{long2018conditional} and ToAlign~\cite{wei2021metaalign}. They leverage information from both source and target domains to reduce distribution discrepancy. 
(3) Unsupervised graph domain adaptation method, e.g., CoCo~\cite{yin2023coco}, which is the state-of-the-art source-free domain adaptation method designed for image classification.

\textbf{Implementation Details.}~We employ a 3-layer GNN encoder (GIN by default) with an embedding dimension of 64.  The model is optimized using Adam optimizer with a learning rate of 0.001 and a dropout rate of 0.3.  We train the model for 200 epochs with a temperature parameter of 0.1 for contrastive learning. For baselines, we configure the methods with the same hyperparameters from their original papers and further fine-tune them to optimize performance.  All experiments are conducted with PyTorch on  NVIDIA A100-SXM4-80GB. To reduce randomness, we perform 5 runs with different random seeds and report the average accuracy.

\setlength{\tabcolsep}{0.3em}
\begin{table}[t]
    \centering
    \caption{The results (in \%) on Mutagenicity (source→target). The \textcolor{red}{red} and \textcolor{blue}{blue} numbers denote the highest and second highest results.}
    \resizebox{\textwidth}{21mm}{
    \begin{tabular}{@{}l|cccccccccccc|c@{}}
    \toprule
    Methods & M0-M1 & M1-M0 & M0-M2 & M2-M0 & M0-M3 & M3-M0 & M1-M2 & M2-M1 & M1-M3 & M3-M1 & M2-M3 & M3-M2 & Avg. \\ \midrule
    GCN     & 73.5 & 60.8 & 69.6 & 68.5 & 54.2 & 55.1 & 68.6 & 75.3 & 51.4 & 46.2 & 58.6 & 60.1 & 61.8 \\
    GIN     & \textcolor{blue}{77.3} & 68.9 & 70.2 & 69.1 & 63.8 & 61.8 & \textcolor{blue}{77.6} & \textcolor{blue}{78.3} & 64.2 & 71.5 & 69.2 & \textcolor{blue}{72.8} & 70.4 \\
    GMT     & 67.2 & 52.3 & 59.8 & 47.5 & 53.2 & 52.5 & 59.8 & 67.3 & 46.7 & 67.2 & 53.1 & 59.8 & 57.2 \\
    GAT     & 65.5 & 71.3 & 57.6 & 63.1 & 38.2 & 51.6 & 59.7 & 58.8 & \textcolor{blue}{72.7} & 57.6 & \textcolor{red}{79.5} & 67.6 & 61.9 \\
    GraphSAGE & 69.3 & 69.2 & 60.4 & 63.7 & 42.3 & 56.5 & 62.4 & 61.6 & 70.5 & 58.3 & \textcolor{blue}{79.1} & 65.3 & 63.2 \\
    DeSGDA  & 76.5 & \textcolor{blue}{72.3} & \textcolor{blue}{71.4} & \textcolor{blue}{71.5} & 60.7 & \textcolor{blue}{67.3} & 75.2 & \textcolor{red}{79.4} & 62.7 & \textcolor{red}{75.8} & 65.5 & 72.1 & \textcolor{blue}{70.9} \\\midrule
    CDAN    & 75.3 & 71.2 & 70.7 & 70.3 & 58.7 & 58.4 & 70.1 & 76.1 & 58.3 & 69.4 & 58.7 & 63.5 & 66.7 \\
    ToAlign & 67.3 & 47.5 & 59.8 & 47.5 & 46.7 & 47.2 & 59.6 & 67.2 & 46.7 & 67.3 & 46.5 & 59.8 & 55.3 \\
    CoCo    & 75.4 & 71.7 & 68.7 & 69.2 & \textcolor{blue}{60.8} & 65.7 & \textcolor{red}{79.2} & 76.8 & 63.4 & \textcolor{blue}{73.6} & 64.6 & 70.1 & 69.9 \\ \midrule
    FracNet & \textcolor{red}{79.4} & \textcolor{red}{76.3} & \textcolor{red}{72.8} & \textcolor{red}{73.4} & \textcolor{red}{73.9} & \textcolor{red}{74.1} & 73.5 & 77.8 & \textcolor{red}{74.0} & 72.1 & 74.0 & \textcolor{red}{73.4} & \textcolor{red}{74.6} \\ \bottomrule
    \end{tabular}}
\end{table}

\setlength{\tabcolsep}{0.4em}
\begin{table}[ht]
    \centering
    \caption{The results (in \%) on NCI1 (source→target). The \textcolor{red}{red} and \textcolor{blue}{blue} numbers denote the highest and second highest results.}
    \resizebox{\textwidth}{21mm}{
    \begin{tabular}{@{}l|cccccccccccc|c@{}}
    \toprule
    Methods & N0-N1 & N1-N0 & N0-N2 & N2-N0 & N0-N3 & N3-N0 & N1-N2 & N2-N1 & N1-N3 & N3-N1 & N2-N3 & N3-N2 & Avg. \\ \midrule
    GCN     & 51.2 & 70.2 & 42.7 & 27.6 & 32.1 & 27.1 & 55.2 & 50.6 & 50.9 & 49.1 & 67.3 & 57.5 & 48.5 \\
    GIN     & 66.8 & 78.4 & 60.2 & 72.3 & 51.1 & 68.6 & 63.5 & 67.8 & 65.9 & 60.3 & 71.1 & 67.2 & 66.1 \\
    GMT     & 50.6 & 72.9 & 57.3 & 72.8 & 66.4 & 73.1 & \textcolor{blue}{72.4} & 50.8 & 66.5 & 58.3 & 66.3 & \textcolor{red}{72.6} & 65.0 \\
    GAT     & 67.2 & 62.5 & 63.6 & 70.1 & 61.4 & 59.7 & 63.9 & 68.5 & 66.3 & \textcolor{blue}{64.9} & 64.6 & 68.1 & 65.1 \\
    GraphSAGE & 67.5 & 70.6 & 61.3 & 69.2 & 65.8 & 64.7 & 68.5 & 66.2 & 64.2 & 59.4 & 63.9 & 68.4 & 65.8 \\
    DeSGDA  & 64.4 & 76.9 & \textcolor{blue}{64.8} & 76.1 & \textcolor{blue}{68.6} & \textcolor{red}{74.1} & 66.8 & 64.6 & \textcolor{blue}{69.2} & 63.8 & 70.5 & 64.2 & 68.7 \\\midrule
    CDAN    & 57.1 & 74.7 & 61.2 & 73.7 & 68.2 & 73.3 & 60.2 & 56.5 & 68.2 & 53.9 & 68.4 & 59.6 & 64.6 \\
    ToAlign & 49.1 & 27.2 & 57.3 & 27.1 & 66.4 & 27.1 & 57.2 & 49.1 & 66.4 & 49.1 & 66.5 & 57.3 & 50.0 \\
    CoCo    & \textcolor{blue}{69.7} & \textcolor{red}{80.2} & {64.5} & \textcolor{red}{76.3} & {64.6} & \textcolor{blue}{73.8} & {68.2} & \textcolor{blue}{70.2} & {67.7} & {61.2} & \textcolor{blue}{73.1} & 64.8 & \textcolor{blue}{69.5} \\ \midrule
    FracNet & \textcolor{red}{74.1} & \textcolor{blue}{75.6} & \textcolor{red}{72.4} & \textcolor{blue}{74.2} & \textcolor{red}{69.3} &{68.9} & \textcolor{red}{74.1} & \textcolor{red}{73.9} & \textcolor{red}{74.0} & \textcolor{red}{69.5} & \textcolor{red}{74.1} & \textcolor{blue}{71.2} & \textcolor{red}{72.6} \\ \bottomrule
    \end{tabular}}
\end{table}

\setlength{\tabcolsep}{0.4em}
\begin{table}[ht]
    \centering
    \caption{The results (in \%) on PROTEINS (source→target). The \textcolor{red}{red} and \textcolor{blue}{blue} numbers denote the highest and second highest results.}
    \resizebox{\textwidth}{21mm}{
    \begin{tabular}{@{}l|cccccccccccc|c@{}}
    \toprule
    Methods & P0-P1 & P1-P0 & P0-P2 & P2-P0 & P0-P3 & P3-P0 & P1-P2 & P2-P1 & P1-P3 & P3-P1 & P2-P3 & P3-P2 & Avg. \\ \midrule
    GCN     & 73.7 & 82.6 & 57.5 & 83.9 & 24.4 & 17.3 & 57.6 & 70.8 & 24.5 & 26.3 & 37.5 & 42.4 & 49.9 \\
    GIN     & 71.6 & 70.2 & 58.4 & 56.9 & 74.2 & 78.2 & 63.3 & 67.1 & 35.8 & 60.8 & 71.6 & 65.2 & 64.4 \\
    GMT     & 73.6 & 82.5 & 57.6 & 83.1 & \textcolor{blue}{75.6} & 17.3 & 57.6 & 73.5 & 75.4 & 26.3 & {75.5} & 42.3 & 61.7 \\
    GAT     & 67.2 & 66.3 & 68.5 & 71.4 & 70.6 & 53.5 & 65.1 & 64.6 & 58.2 & 57.5 & 70.9 & 68.1 & 65.2 \\
    GraphSAGE & 70.5 & 66.8 & 66.4 & 72.3 & 71.7 & 63.7 & 64.7 & 68.1 & 59.6 & 60.1 & 71.6 & 69.2 & 67.1 \\
    DeSGDA  & \textcolor{red}{77.5} & \textcolor{blue}{84.3} & \textcolor{blue}{70.2} & \textcolor{blue}{84.2} & \textcolor{red}{76.6} & \textcolor{red}{83.2} & \textcolor{blue}{71.6} & \textcolor{blue}{77.2} & \textcolor{blue}{75.8} & {73.4} & 75.4 & \textcolor{blue}{70.4} & \textcolor{blue}{76.6} \\ \midrule
    CDAN    & 75.8 & 83.1 & 60.6 & 82.6 & 75.8 & 70.5 & 64.7 & \textcolor{red}{77.4} & 73.1 & \textcolor{red}{75.4} & \textcolor{blue}{75.6} & 67.1 & 73.5 \\
    ToAlign & 73.2 & 82.5 & 57.4 & 82.3 & 24.3 & {82.6} & 57.5 & 73.7 & 24.3 & 73.6 & 24.2 & 57.6 & 59.4 \\
    CoCo    & 74.6 & 83.9 & 65.2 & 83.4 & 72.1 & \textcolor{blue}{82.7} & 69.5 & {75.4} & 70.7 & 73.2 & 72.4 & 66.1 & 74.1 \\ \midrule
    FracNet & \textcolor{blue}{76.4} & \textcolor{red}{87.5} & \textcolor{red}{70.7} & \textcolor{red}{87.9} & 74.7 & 72.4 & \textcolor{red}{74.5} & 74.3 & \textcolor{red}{77.2} & \textcolor{blue}{73.7} & \textcolor{red}{77.3} & \textcolor{red}{73.7} & \textcolor{red}{76.9} \\ \bottomrule
    \end{tabular}}
\end{table}

\subsection{Performance Comparison (RQ1)}
We conduct extensive experiments across three benchmark datasets (Mutagenicity, NCI1, and PROTEINS) to evaluate FracNet against state-of-the-art baselines. 
The comprehensive results in Tables 1-3 demonstrate consistent superiority of our approach.

Traditional GNN methods (GCN, GIN, GAT, etc.) often struggle with domain adaptation tasks, particularly in challenging scenarios like Mutagenicity M0→M3 (54.2\%) and PROTEINS P0→P3 (24.4\%), where performance drops substantially. These methods fail to account for structural and distributional shifts between domains, leading to suboptimal transfer learning.

Domain adaptation approaches show improved but inconsistent performance. While DeSGDA achieves competitive results on PROTEINS (76.6\% average accuracy), its effectiveness varies considerably across datasets (70.9\% on Mutagenicity and 68.7\% on NCI1). Similarly, CoCo performs well on certain transfer tasks (e.g., 80.2\% on NCI1 N1→N0) but lacks robustness across broader evaluation settings. This instability stems from their monolithic treatment of graph representations, which fails to address the frequency-dependent nature of domain shifts.

FracNet demonstrates consistent state-of-the-art performance across all datasets, achieving the highest average accuracy on Mutagenicity (74.6\%), NCI1 (72.6\%), and PROTEINS (76.9\%). The performance advantage is particularly pronounced in challenging transfer scenarios such as Mutagenicity M0→M3 (73.9\% vs. next best 64.6\%) and NCI1 N1→N3 (74.0\% vs. next best 69.2\%), where domain shifts are most severe. The significant improvement over strong baselines (+2.3\% over DeSGDA on PROTEINS, +4.1\% over CoCo on Mutagenicity, and +3.1\% over CoCo on NCI1) validates the effectiveness of our approach.

\subsection{Case Study (RQ2,RQ3)}

\subsubsection{4.3.1 Spectral Shift Direction and Distance (RQ2)}

\begin{figure}[ht]
\vspace{0.0005cm}
    \centering
    \includegraphics[width=1\textwidth]{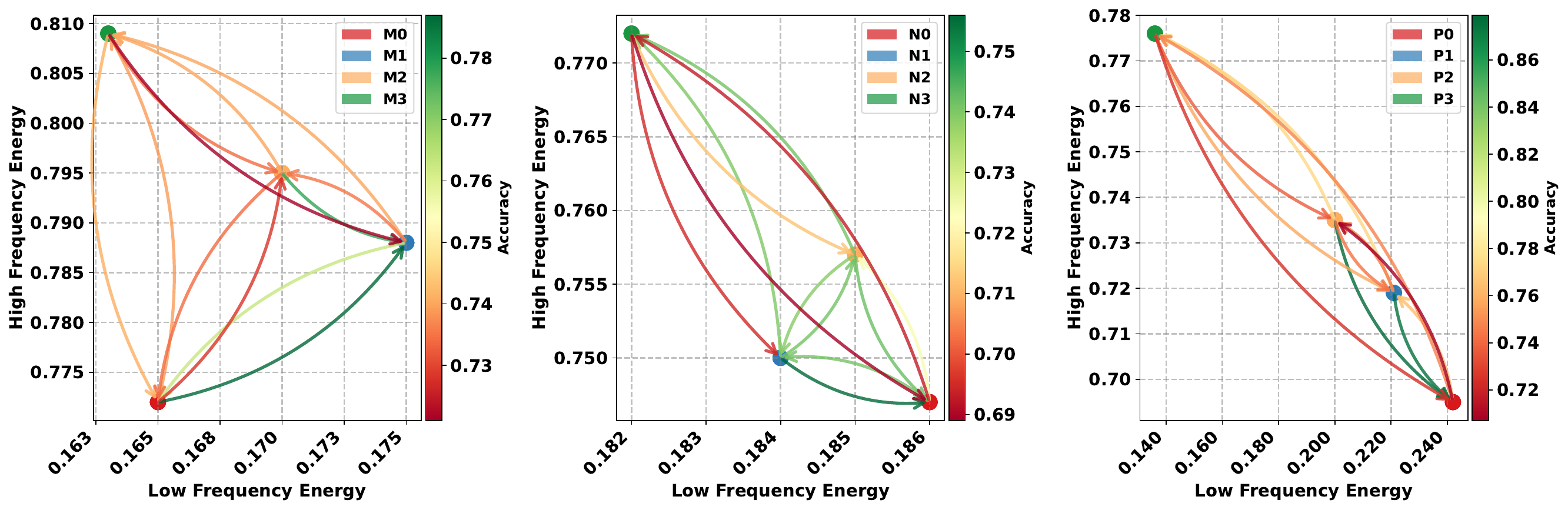}
    \caption{Energy trajectories in frequency domain on three datasets.}
    \label{fig:casetop}
\end{figure}

We analyzed spectral properties of molecular domains by decomposing graph Laplacian eigenvalues into low and high-frequency components as illustrated in Figure~\ref{fig:casetop}, where x-axis and y-axis represent the low-frequency energy and high-frequency energy for different domains, respectively.

Figure~\ref{fig:casetop} maps domains in spectral energy space, revealing fundamental patterns governing transfer learning effectiveness. The \textit{spectral distance} between domains influences adaptation quality, with the one of moderate distance typically outperforming the one of larger distance across all datasets.
More importantly, we observe a consistent \textit{directional asymmetry} in transfer performance: when high-frequency energy remains relatively low, knowledge transfer from domains with lower to higher low-frequency energy achieves superior performance (e.g., N1→N0 outperforms N0→N1). This establishes a key principle that when the high-frequency energy is low, adaptation along \textbf{ascending} low-frequency energy gradients facilitates more effective adaption between molecular domains.

\begin{figure}[ht]
\vspace{0.0005cm}
    \centering
    \includegraphics[width=1\textwidth]{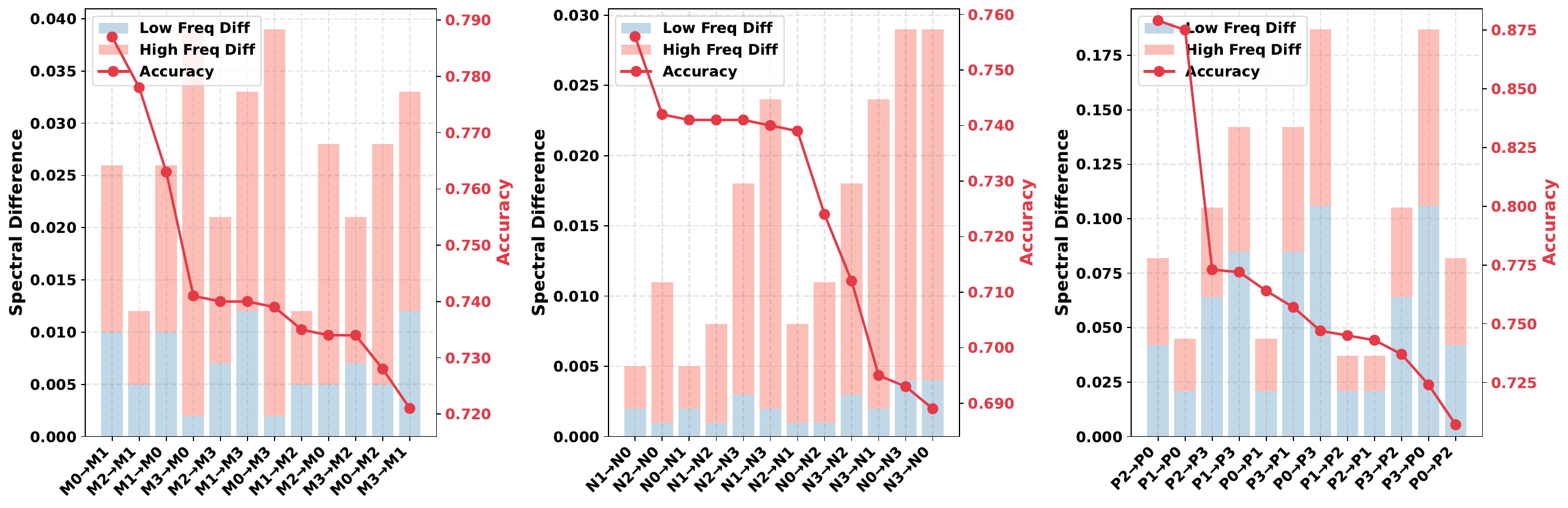}
    \caption{Pairwise spectral differences, decomposed into low-frequency (blue) and high-frequency (pink) components.}
    \label{fig:caselow}
\end{figure}
\subsubsection{4.3.2 Proportion of Different Frequency Components (RQ2)}

% 同样，先介绍下我们怎么得到那个bar chart的
To further quantify spectral differences between molecular domains, we calculated the normalized energy distribution across frequency bands for each domain pair. For each transfer task, we decomposed the spectral energy difference into low-frequency and high-frequency components, visualizing their relative contributions as stacked bar charts, as shown in Figure~\ref{fig:caselow}. Each rectangle is the difference between every two domains in terms of low-frequency (blue part) or high-frequency (pink part) regime, with a larger size indicating a larger difference.

Notably, Figure~\ref{fig:caselow} presents a distinctive pattern that explains FracNet's relatively smaller performance advantage on PROTEINS compared to other datasets. 
Unlike Mutagenicity and NCI1, PROTEINS transfer tasks exhibit substantially more balanced distributions between high and low-frequency differences, as evidenced by the more equal heights of pink and blue bars. Many PROTEINS transfer tasks show significant low-frequency contributions (larger blue portions).
This balanced frequency profile in PROTEINS creates a unique challenge: when frequency differences are more evenly distributed across bands, conventional domain adaptation methods can partially compensate through their unified representation approach. The advantage of frequency-specific processing becomes less pronounced in such scenarios, explaining why FracNet shows a smaller margin of improvement (0.3\% over DeSGDA) on PROTEINS compared to Mutagenicity (3.7\% over CoCo) and NCI1 (3.1\% over CoCo).

\subsubsection{4.3.3 Graph structure analysis (RQ3)}

Beyond spectral analysis, we examine graph structural properties to better understand how FracNet works. We chose cyclomatic numbers (count of independent cycles in a graph) as our structural metric because they directly influence spectral properties. Cycles create distinctive patterns in the Laplacian eigenvalue spectrum by introducing closed paths that alter graph connectivity structures. The distribution and density of cycles significantly shape the spectral energy profile: graphs with more cycles typically exhibit different eigenvalue distributions compared to sparser structures. This cycle distribution directly impacts how energy spreads across frequency bands in the graph spectrum, providing a concrete structural interpretation of our spectral observations.

Our cyclomatic distribution analysis reveals clear relationships between graph structural complexity and transfer performance. PROTEINS exhibits extreme cyclomatic variation (\textbf{maximum 539 cycles}) compared to Mutagenicity and NCI1 (\textbf{maxima of 16 and 18 cycles}). This structural disparity correlates with FracNet's performance patterns. While achieving high absolute accuracy (76.9\%) on PROTEINS, FracNet shows minimal relative improvement (0.3\% over DeSGDA). This suggests that excessive cycle distribution differences create spectral profiles too disparate for effective frequency-adaptive processing.

FracNet demonstrates its strongest performance advantages on Mutagenicity (3.7\% over DeSGDA) and NCI1 (3.1\% over CoCo), both characterized by moderate cyclomatic variations. These results identify an optimal efficacy zone for spectral adaptation methods where structural differences are significant enough to benefit from frequency-specific processing but not so extreme as to create fundamentally incompatible spectral distributions. 

\begin{figure}[ht]
\vspace{0.0005cm}
    \centering
    \includegraphics[width=0.8\textwidth]{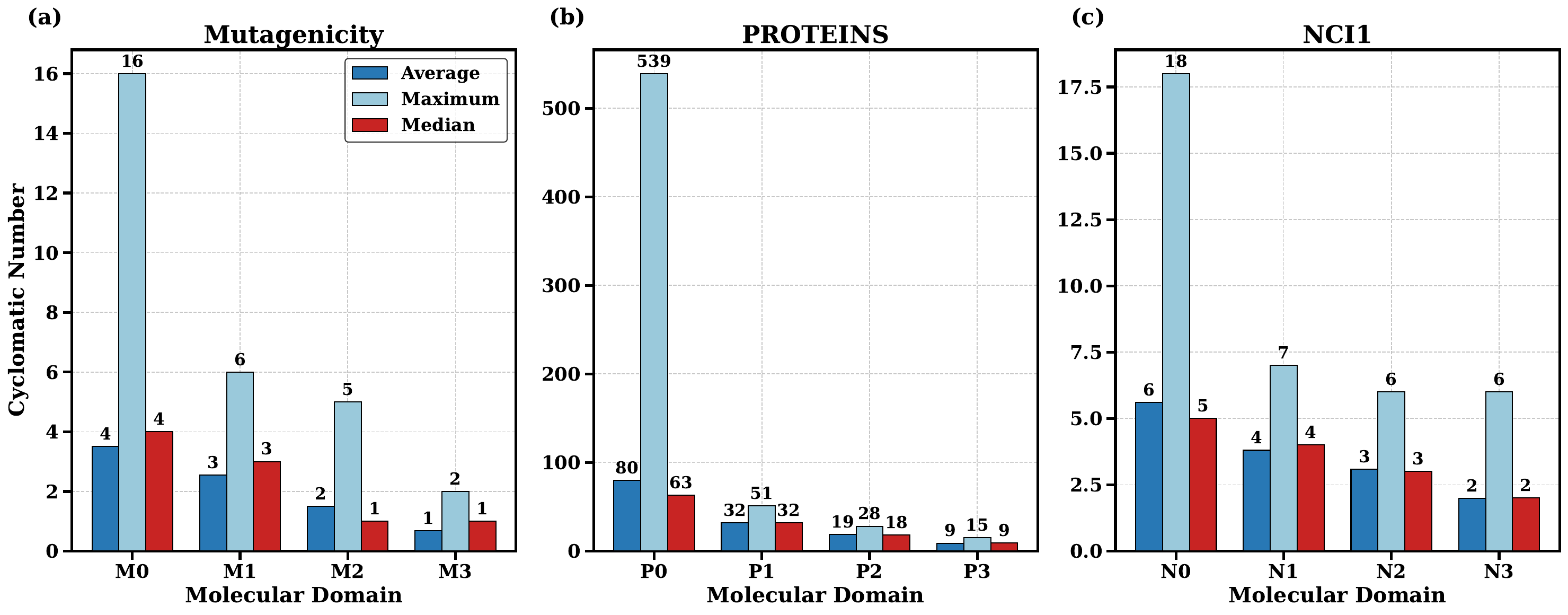}
    \caption{Cyclomatic number distributions across molecular domains in three datasets. }
    \label{fig:cyclo}
\end{figure}

\subsection{Ablation Study (RQ4)}

%\begin{figure}[t!]
%    \centering
%    \includegraphics[width=0.5\textwidth]{figures/abl.png}
%    \caption{Ablation study results showing average accuracy on (a) Mutagenicity and (b) PROTEINS.}
%   \label{fig:abla}
%\end{figure}

\begin{figure}[t!]
    \centering
    \subfigure[Temperature coefficient ($\tau$) and balance parameter ($\gamma$). (The upper two represent the Mutagenicity dataset and the lower two the NCI1 dataset.)]{
        \includegraphics[width=0.48\textwidth]{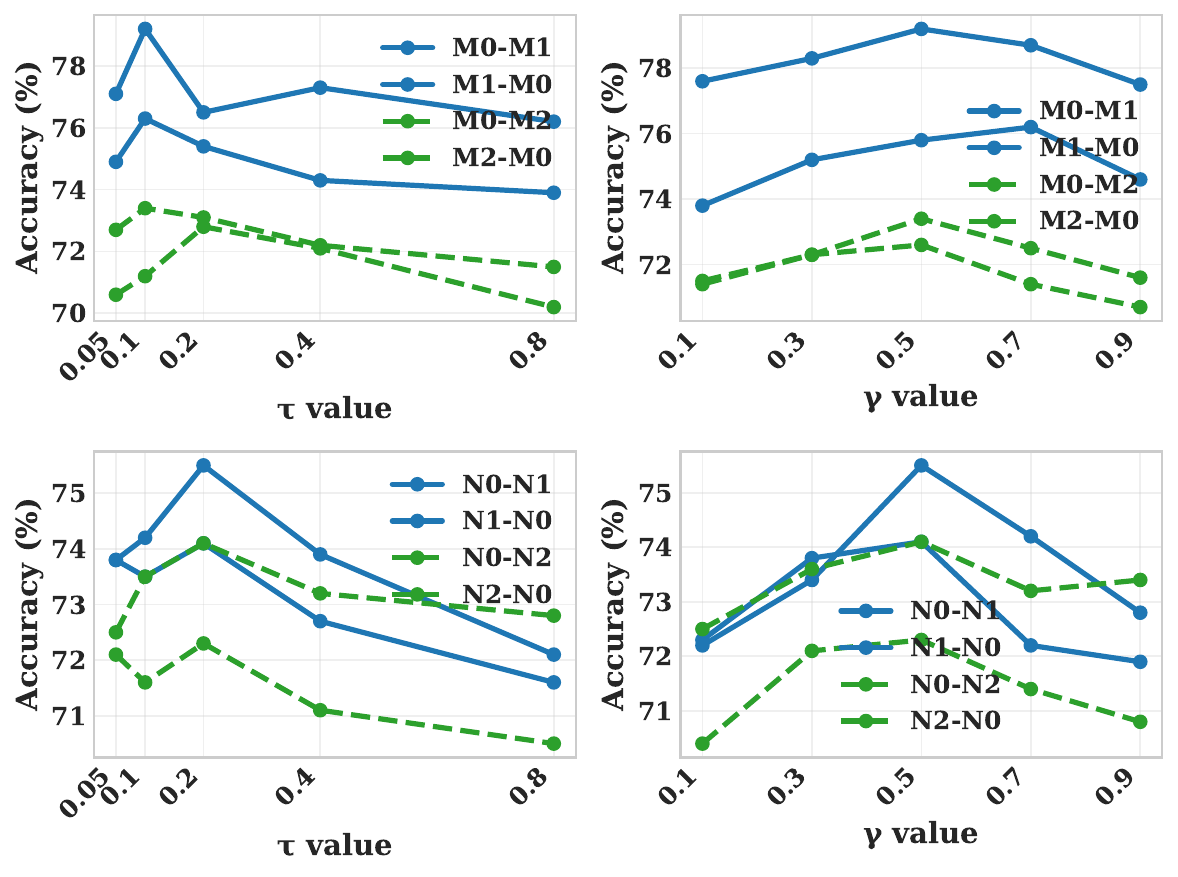}
        \label{fig:sensitivity}
    }
    \hfill % 水平间距
    \subfigure[Ablation study results showing average accuracy on (a) Mutagenicity and (b) PROTEINS.]{
        \includegraphics[width=0.48\textwidth]{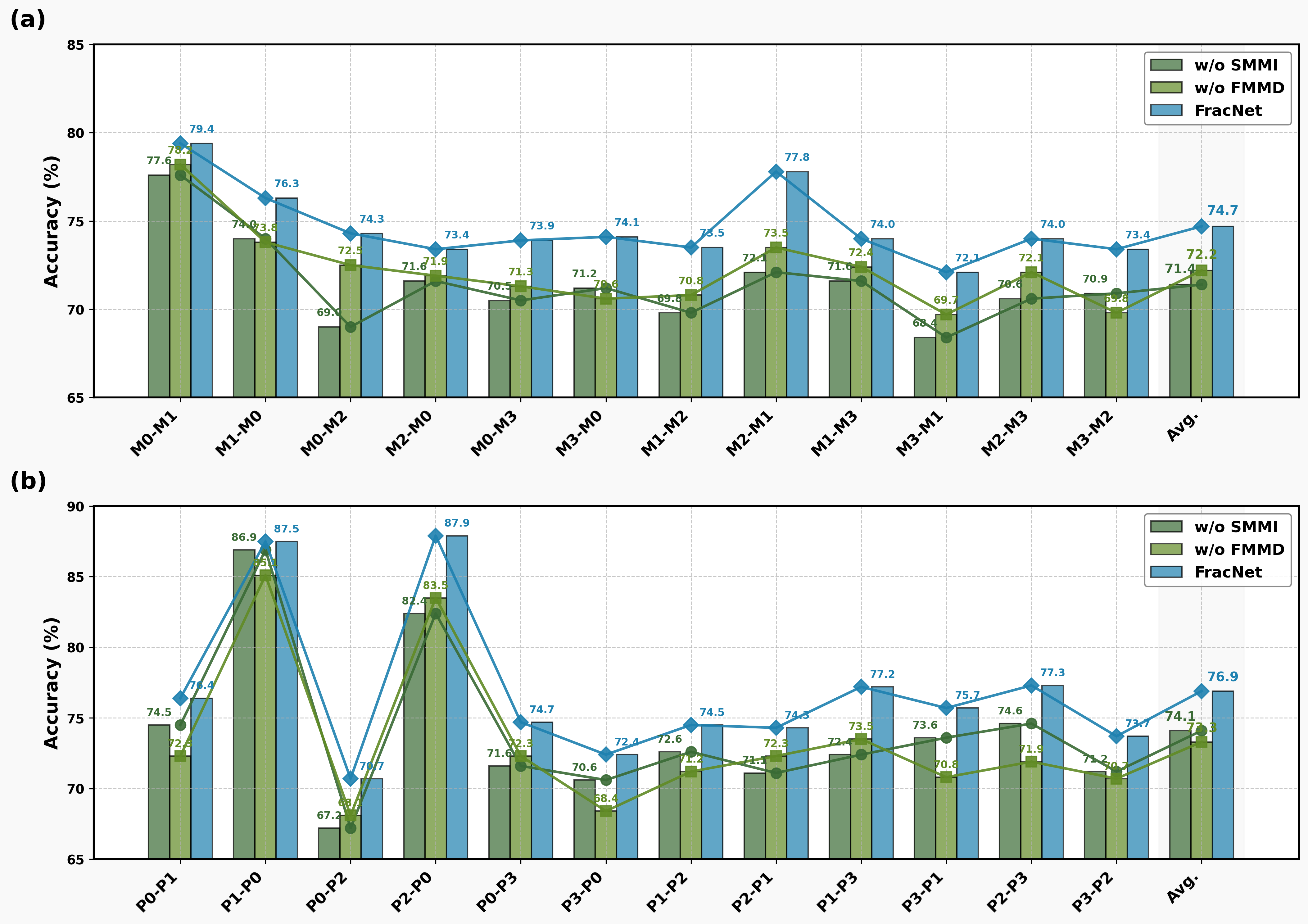}
        \label{fig:abla}
    }
    \caption{(a) Hyperparameter sensitivity analysis; (b) Ablation study results.}
    \label{fig:combined}
\end{figure}
We evaluate each component's contribution through ablation studies on two datasets by creating variants: (1) \textbf{w/o SMMI}, removing the Spectral-guided Maximum Mutual Information module; and (2) \textbf{w/o FMMD}, removing the Frequency-aware Maximum Mean Discrepancy module. 

As shown in Figure~\ref{fig:abla}, both components significantly enhance performance. On Mutagenicity, removing SMMI and FMMD reduces accuracy by 3.3\% and 2.5\% respectively, with similar patterns on PROTEINS (2.8\% and 3.6\% decreases).
SMMI proves crucial for challenging transfers with significant domain shifts (M0→M2, P0→P2) by maintaining discriminative boundaries while aligning domains. FMMD contributes most to transfers involving structurally complex domains (M2→M1, P2→P0), validating its adaptive frequency-aware design for capturing domain-specific properties across frequency bands.

Interestingly, component contributions vary by dataset. In PROTEINS, with balanced high-low frequency differences, FMMD contributes more significantly. In Mutagenicity, where high-frequency differences dominate, SMMI plays a more critical role. These patterns confirm our spectral analysis: FMMD's adaptive frequency handling excels with balanced spectral differences, while SMMI's discriminative preservation becomes vital when high-frequency components dominate domain shifts. Together, these complementary components address domain adaptation's dual challenges: preserving discriminative information while effectively aligning domains across the spectral dimension.

%\begin{figure}[t!]
%    \centering
%    \includegraphics[width=0.48\textwidth]{figures/hyperparameter_sensitivity.pdf}
%    \caption{Temperature coefficient ($\tau$) and balance parameter ($\gamma$). (The upper two represent the Mutagenicity dataset and the lower two the NCI1 dataset.)
%    }
%    \label{fig:sensitivity}
%\end{figure}

% 后面的文字继续

\subsection{Sensitivity Analysis}

The sensitivity analysis of FracNet's critical hyperparameters reveals insightful patterns regarding the model's robustness and optimal configuration across different molecular domains (Figure~\ref{fig:sensitivity}). As shown in subfigures (a) and (c), the temperature parameter $\tau$, which governs the spectral filtering sharpness, demonstrates a consistent bell-shaped performance curve with optimal values centered around $\tau=0.1\sim0.2$ across both Mutagenicity and NCI1 datasets. This moderate filtering threshold achieves an optimal balance between preserving essential structural information and eliminating domain-specific noise. Notably, performance degradation is more pronounced for larger $\tau$ values ($\tau>0.4$), with accuracy dropping by up to 3.0\% in M0→M1 and 2.5\% in N1→N0 transfers, suggesting that excessive spectral compression eliminates crucial molecular structural signals. Similarly, the frequency modulation parameter $\gamma$, examined in subfigures (b) and (d), exhibits optimal performance at $\gamma=0.5$ across most transfer tasks, with a more gradual performance decline toward extreme values. 
\section{Conclusion}
We have presented FracNet with two synergic modules to decompose the original graph into high-frequency and low-frequency components and perform frequency-aware domain adaptation. The key insight is that domain shifts can be better understood through spectral analysis, where low-frequency components encode domain-invariant global patterns, and high-frequency components capture domain-specific local details. Moreover, the blurring boundary problem of domain adaptation is improved by integrating with a contrastive learning framework. Besides providing rigorous theoretical proof, we conducted extensive experiments across three benchmark datasets to demonstrate the significant performances of FracNet.  Future work includes extending FracNet to multi-source domain adaptation scenarios and exploring applications in other graph domain adaptation tasks.

\section{Acknowledgments}
Wenqi FAN is partly supported by General Research Funds from the Hong Kong Research Grants Council (project no. PolyU 15207322, 15200023, 15206024, and 15224524), internal research funds from The Hong Kong Polytechnic University (project no. P0042693, P0048625, P0051361, P0052406, and P0052986). 

\end{document}